\documentclass[conference]{IEEEtran}
\IEEEoverridecommandlockouts

\usepackage{amsmath,amsfonts,bm}

\def\eqref#1{equation~\ref{#1}}

\def\1{\bm{1}}

\DeclareMathAlphabet{\mathsfit}{\encodingdefault}{\sfdefault}{m}{sl}
\SetMathAlphabet{\mathsfit}{bold}{\encodingdefault}{\sfdefault}{bx}{n}

\usepackage{amssymb,graphicx,booktabs}
\usepackage{cite,flafter}
\usepackage{url}

\title{VGGT-VLA-World: Intent-Conditioned 3D World Evolution for Autonomous Driving
\thanks{\textsuperscript{*}Corresponding authors: Changye Liu, Ziying Song, and Diange Yang.}}
\author{
\IEEEauthorblockN{Zhaoyang Liu}
\IEEEauthorblockA{\textit{Tsinghua University}\\
Beijing, China\\
\url{liuzhaoy24@mails.tsinghua.edu.cn}}
\and
\IEEEauthorblockN{Zhigui Lin}
\IEEEauthorblockA{\parbox[t]{0.22\textwidth}{\centering\footnotesize
\textit{SAIC GM Wuling Automobile Co.,Ltd.}\\
\textit{Guangxi Laboratory of New Energy Automobile}\\
\textit{Guangxi Key Laboratory of Automobile Artificial Intelligence}\\
Guangxi, China\\
\url{zhigui.lin@sgmw.com.cn}}}
\and
\IEEEauthorblockN{Kun Jiang}
\IEEEauthorblockA{\textit{Tsinghua University}\\
Beijing, China\\
\url{jiangkun@tsinghua.edu.cn}}
\and
\IEEEauthorblockN{Tianhui Li}
\IEEEauthorblockA{\parbox[t]{0.22\textwidth}{\centering\footnotesize
\textit{SAIC GM Wuling Automobile Co.,Ltd.}\\
\textit{Guangxi Laboratory of New Energy Automobile}\\
\textit{Guangxi Key Laboratory of Automobile Artificial Intelligence}\\
Guangxi, China\\
\url{tianhui.li@sgmw.com.cn}}}
\and[\hfill\break\mbox{}\hfill]
\IEEEauthorblockN{Changye Liu\textsuperscript{*}}
\IEEEauthorblockA{\parbox[t]{0.22\textwidth}{\centering\footnotesize
\textit{SAIC GM Wuling Automobile Co.,Ltd.}\\
\textit{Guangxi Laboratory of New Energy Automobile}\\
\textit{Guangxi Key Laboratory of Automobile Artificial Intelligence}\\
Guangxi, China\\
\url{changye.liu@sgmw.com.cn}}}
\and
\IEEEauthorblockN{Ziying Song\textsuperscript{*}}
\IEEEauthorblockA{\textit{Nanyang Technological University}\\
Singapore, Singapore\\
\url{22110110@bjtu.edu.cn}
\mbox{}}
\and
\IEEEauthorblockN{Diange Yang\textsuperscript{*}}
\IEEEauthorblockA{\textit{Tsinghua University}\\
Beijing, China\\
\url{ydg@mail.tsinghua.edu.cn}}
}

\begin{document}

\maketitle

\begin{figure*}[t]
    \centering
    \includegraphics[width=\textwidth]{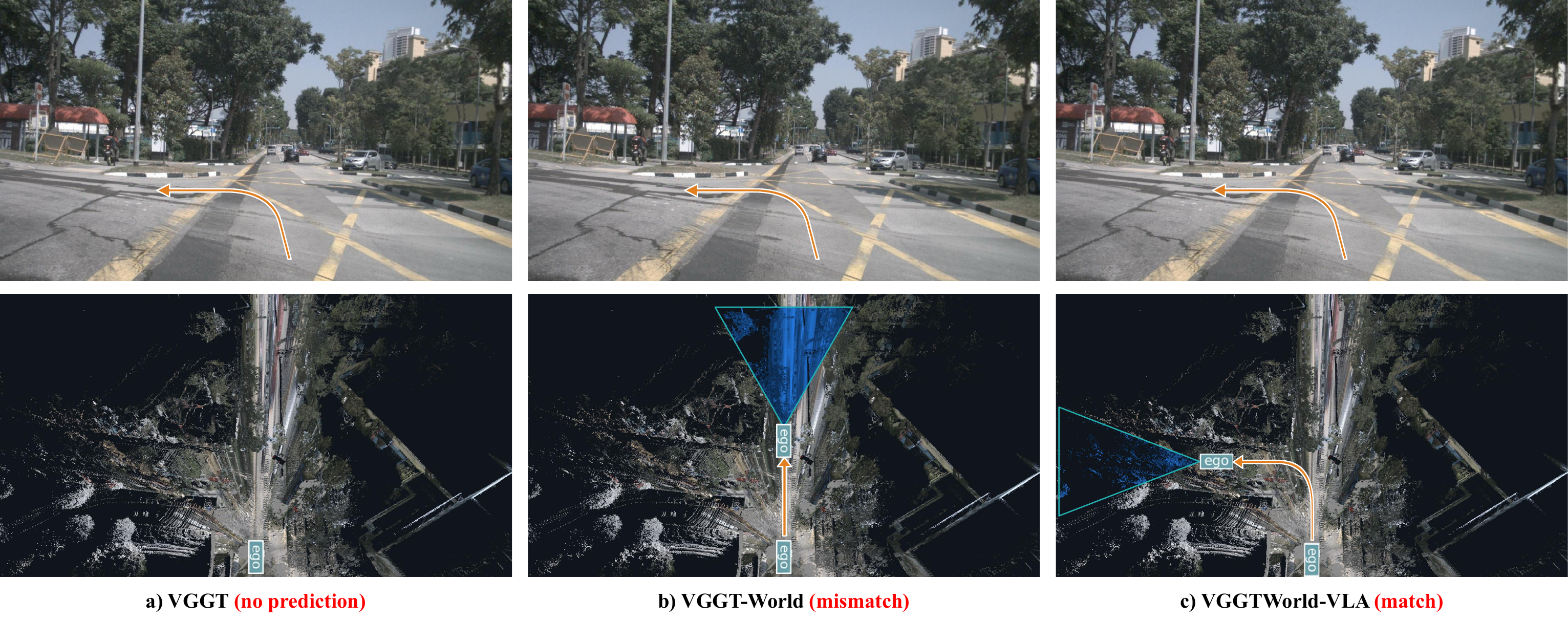}
    \caption{
\textbf{Motivation for intention-conditioned 3D world prediction.}
Geometric reconstruction captures the observed scene,
while geometry-based world models extend this capability
to future prediction, primarily from historical observations.
Without explicit intention and action conditioning,
the forecast may depict straight driving despite an
intended left turn.
Our approach incorporates driving semantics and ego actions
to predict future geometry consistent with the intended maneuver.
    }
    \label{fig:motivation}
\end{figure*}

\begin{abstract}
VGGT provides a strong foundation for geometry-centric world models by recovering unified 3D scene geometry from visual observations. Although recent extensions enable temporal 3D prediction, their future evolution remains weakly conditioned on driving intentions and actions, limiting their ability to model alternative action-dependent futures.
We propose \textbf{VGGTWorld-VLA}, an intention-conditioned extension of VGGT-World for controllable 3D world evolution in autonomous driving. First, we introduce an action--semantic conditioning mechanism that injects complementary driving semantics and ego-motion representations into the future-token stream, enabling different future geometry predictions for the same observed scene under alternative ego actions. Second, we develop a geometry--language--action bridge that adapts historical geometry, VLA semantic features, and maneuver and trajectory representations for joint conditioning of future geometry prediction.
We evaluate future geometry prediction on NAVSIM, with ablation studies examining the contributions of semantic and action conditioning. Compared with baseline, our method reduces future-depth RMSE by 13.7\% and improves $\delta_1$ by 4.11 percentage points on NAVSIM motion subset. Controlled action-input interventions further show that the model uses the supplied action information. These results support semantic and action conditioning for controllable VGGT-based world prediction in autonomous driving.
\end{abstract}


\section{Introduction}

Recovering 3D scene geometry from visual observations is fundamental to
autonomous driving, where decisions depend on depth, spatial structure,
and scene motion~\cite{zuo2026dvgt}. Visual Geometry Grounded Transformer
(VGGT)~\cite{wang2025vggt} unifies camera estimation, depth prediction,
point-map reconstruction, and point tracking in a feed-forward model. Its strong generalization and unified geometry tokens provide a
promising foundation for representing driving scenes. More importantly,
these representations can potentially support not only reconstruction
of the observed scene, but also prediction of its future 3D evolution~\cite{sun2026vggtworld}.

Following VGGT, several studies have adapted geometry foundation models
to autonomous driving. DriveVGGT~\cite{jia2025drivevggt} and
DVGT~\cite{zuo2026dvgt} improve metric reconstruction for multi-camera
driving scenes, while DynamicVGGT~\cite{he2026dynamicvggt} extends
geometry estimation to dynamic 4D reconstruction. VGGDrive
~\cite{wang2026vggdrive} injects VGGT features into a driving VLM, and
DVGT-2~\cite{zuo2026dvgt2} jointly models dense geometry and trajectory
planning. These developments highlight the potential of VGGT-based
representations for driving scene understanding, providing a foundation
for extending geometry reconstruction toward future world prediction.

VGGT-World~\cite{sun2026vggtworld} extends VGGT from static reconstruction to temporal world modeling by predicting the evolution of frozen geometry tokens with an autoregressive flow transformer, enabling efficient and geometrically consistent future 3D prediction. Nevertheless, its future evolution is primarily determined by historical observations and is not explicitly conditioned on driving intentions or actions~\cite{sun2026vggtworld}. It therefore cannot adequately distinguish the different physical futures induced by alternative behaviors, such as lane keeping, lane changing, or yielding. Existing geometry-aware driving models, including VGGDrive and DVGT-2, mainly use current geometry for planning rather than evaluating actions through their future 3D consequences~\cite{wang2026vggdrive,zuo2026dvgt2}. High-level semantic representations can provide the missing driving intent, but directly integrating them with VGGT-World is non-trivial because semantic, action, and geometry tokens occupy different representation spaces. The central problem addressed in this work is therefore how to condition geometry-native world evolution on driving semantics and actions, enabling future 3D predictions to explicitly reflect the physical consequences of driving decisions. As illustrated in Fig.~\ref{fig:motivation}, the same observation
can lead to different future viewpoints and 3D observations under
different ego actions. Without explicit action conditioning,
the predicted future may therefore be inconsistent with the
intended driving behavior.

To address this problem, we propose \textbf{VGGTWorld-VLA}, a
VGGT-native world-action modeling framework for autonomous driving.
We use VLA-derived semantic representations together with ego-action
conditions to guide geometry-native world evolution. First, an
\textbf{action--semantic 3D world evolution module} injects
driving semantics and action representations into the future
branch of VGGT-World, enabling the predicted 3D geometry to reflect
action-dependent geometric changes. Second, a
\textbf{geometry--language--action bridge} adapts semantic, action,
and geometry representations for joint conditioning of future
geometry prediction. In this way, the proposed framework connects
driving context and intended ego motion with future 3D prediction
while preserving VGGT-World as the primary world-modeling backbone.

We evaluate VGGTWorld-VLA on NAVSIM~\cite{dauner2024navsim} for future
geometry and depth prediction. Compared with VGGT-World, our method
reduces depth RMSE by 13.7\% and improves $\delta_1$
by 4.11 percentage points, while AbsRel remains comparable
with a modest increase.
Ablation studies further examine the contributions of action
representations and semantic conditioning, while action-input
interventions demonstrate the model's use of scene-aligned
action information.

Our contributions are summarized as follows:
\begin{itemize}
\item We \textbf{extend VGGT-World with semantic and action conditions},
enabling controllable future 3D world evolution.

\item We introduce \textbf{a geometry--language--action bridge} that adapts
driving semantics, multi-representation ego actions, and geometry
features for joint conditioning of future geometry prediction.

\item Experiments on NAVSIM demonstrate \textbf{improved latent prediction
quality and future-depth RMSE and threshold accuracy}, including
13.7\% lower RMSE relative to VGGT-World, supported by
ablation studies and controlled action-input interventions.
\end{itemize}

\section{Related Work}

\subsection{VGGT for Autonomous Driving.}
Visual Geometry Grounded Transformer (VGGT)~\cite{wang2025vggt} unifies camera estimation, depth prediction, point-map reconstruction, and point tracking within a feed-forward architecture, providing a general geometric representation of visual scenes. Its strong generalization has motivated several extensions for autonomous driving. DriveVGGT~\cite{jia2025drivevggt} introduces calibration constraints and cross-camera consistency for metric multi-camera reconstruction, while DVGT~\cite{zuo2026dvgt} recovers globally aligned driving geometry from unposed multi-view images. DynamicVGGT~\cite{he2026dynamicvggt} further models point-level motion for dynamic 4D reconstruction. Beyond perception, VGGDrive~\cite{wang2026vggdrive} injects VGGT features into a driving vision-language model, and DVGT-2~\cite{zuo2026dvgt2} jointly predicts dense geometry and ego trajectories. These studies demonstrate the value of VGGT representations for driving perception and planning, but mainly reconstruct observed scenes or directly use current geometry for action generation, without explicitly reasoning about the action-dependent evolution of the future 3D world.

\subsection{World-Action Models.}
World models predict future states from historical observations,
while world-action models further condition this evolution on
agent actions, enabling alternative futures to be evaluated
before execution~\cite{wang2024drivewm,gao2024vista,li2025wote}. VGGT-World~\cite{sun2026vggtworld} establishes a direct connection between geometry foundation models and world modeling by predicting the temporal evolution of frozen VGGT tokens rather than future RGB frames. This geometry-native formulation offers an efficient representation of future depth and 3D structure, but its prediction remains primarily observation driven. Meanwhile, recent driving models such as DriveVLA-W0~\cite{li2025drivevlaw0}, CoWorld-VLA~\cite{huang2026coworldvla}, and WCog-VLA~\cite{yan2026wcogvla} incorporate future images, world tokens, or agent evolution into action learning. These studies highlight the connection between future world
modeling and action learning~\cite{li2025law,zheng2025world4drive}. Our work focuses on conditioning VGGT-native future geometry prediction on ego actions and driving semantics, with an emphasis on controllable 3D world evolution.

\subsection{Vision-Language-Action Models.}
Vision-language-action models introduce semantic knowledge
and high-level reasoning into autonomous-driving
policies~\cite{pan2024vlp,xu2025drivegpt4v2}. Early studies such as DriveLM~\cite{sima2024drivelm}, LMDrive~\cite{shao2024lmdrive}, and DriveVLM~\cite{tian2024drivevlm} connect visual understanding and language reasoning with driving decisions. More recent methods, including SimLingo~\cite{renz2025simlingo}, OpenDriveVLA~\cite{zhou2025opendrivevla}, AutoVLA~\cite{zhou2026autovla}, and DiffVLA~\cite{jiang2025diffvla}, strengthen language--action alignment, adaptive reasoning, and continuous trajectory generation. Although these models exhibit strong semantic reasoning and planning capabilities, most operate on image or latent visual representations and lack an explicit mechanism for predicting how different intentions and actions alter future 3D geometry. In contrast, our VGGTWorld-VLA uses VLA-derived semantic representations together with given ego-action conditions to control VGGT-based world evolution, connecting semantic reasoning and action-conditioned future 3D prediction within a unified geometry-native framework.

\section{Method}
\label{sec:method}

\begin{figure*}[t]
\centering
\includegraphics[width=\textwidth]{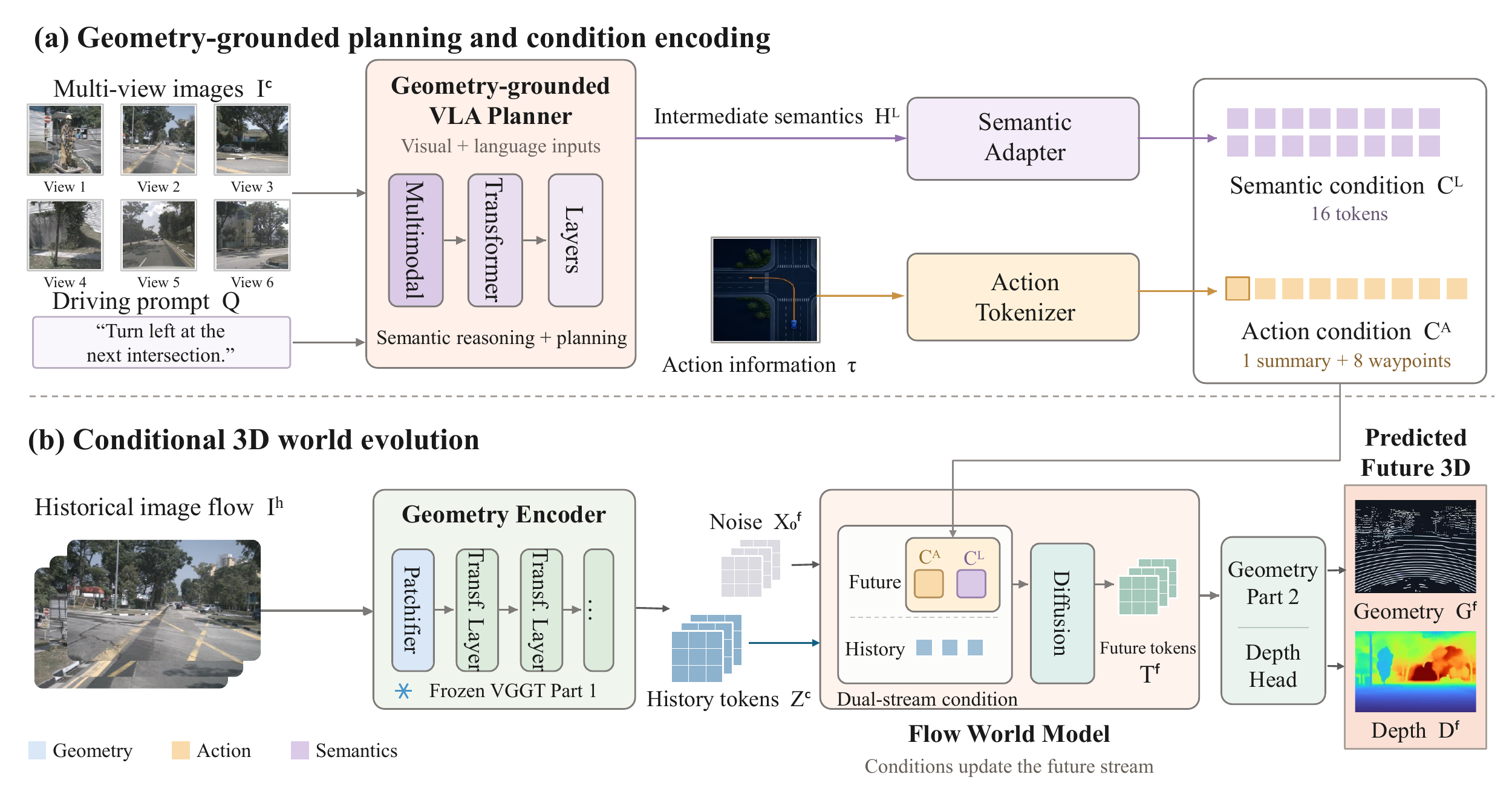}
\caption{
\textbf{Overall architecture of VGGTWorld-VLA.}
The geometric branch encodes historical frames from a
single camera stream, while the semantic branch extracts
VLA features from multi-camera observations and a prompt
at the conditioning timestamp.
A given ego trajectory and its maneuver descriptor are
encoded as action conditions. Action and semantic tokens
condition the future stream of the geometry world model,
whose predictions are decoded into future depth and
3D structure.
}
\label{fig:framework}
\end{figure*}

Fig.~\ref{fig:framework} shows VGGTWorld-VLA,
which predicts future VGGT geometry tokens conditioned
on historical geometry, ego actions, and driving semantics.
A geometry--language--action bridge connects geometry
features from two historical frames, action representations
from a given trajectory and its maneuver descriptor,
and semantic features from multi-camera observations
and a textual prompt.
The predicted tokens are decoded into future depth and
3D geometry using the pretrained VGGT decoder~\cite{wang2025vggt}.
Real future images provide supervision and evaluation
targets only. World prediction is the primary task,
with auxiliary trajectory supervision during joint fine-tuning.

\subsection{Geometry--Language--Action Bridge}
\label{sec:gla_bridge}

The bridge converts historical geometry, ego motion, and
driving semantics into representations that can be consumed
by the world model. The three branches retain their distinct
information sources: geometry describes the observed scene,
action specifies ego motion, and semantic features provide
complementary scene context. Their interaction takes place
within the conditional world model.

\textbf{Historical Geometry Representation.}
Let $\mathbf{I}^{c}$ denote the two historical frames and
$\mathbf{I}^{f}$ the two future frames from the selected
camera stream. The pretrained VGGT encoder~\cite{wang2025vggt}
produces the historical condition and future token target as
\begin{equation}
\mathbf{Z}^{c}
=
E_{\mathrm{G}}(\mathbf{I}^{c}),
\qquad
\mathbf{Z}^{f}
=
\left[
E_{\mathrm{G}}
\left([\mathbf{I}^{c},\mathbf{I}^{f}]\right)
\right]_{f},
\label{eq:geometry_conditions}
\end{equation}
where $[\cdot]_{f}$ selects the future-frame tokens.
The historical condition is independently encoded from
$\mathbf{I}^{c}$ and therefore does not access real future
frames. The full-sequence encoding is used only to construct
future targets. The world model operates on these geometry
tokens rather than directly generating future RGB
images~\cite{sun2026vggtworld}.

\textbf{Multi-Representation Action Encoding.}
Future geometry also depends on both the intended maneuver and
the specific motion of the ego vehicle. A maneuver category
alone cannot distinguish trajectories with different turning
radii or motion speeds. We therefore combine a discrete
maneuver descriptor with continuous trajectory features
to specify the action condition.

Given an ego-centric trajectory
$\boldsymbol{\tau}=\{\mathbf{p}_{i}\}_{i=1}^{N_a}$,
where $\mathbf{p}_{i}=(x_i,y_i)$, we derive a one-hot
maneuver descriptor
$\mathbf{m}=g_{\mathrm{oh}}(\boldsymbol{\tau})$
and represent each waypoint as
\begin{equation}
\mathbf{a}_{i}
=
\left[
\mathbf{p}_{i};
\Delta\mathbf{p}_{i};
i/N_a
\right],
\qquad
\Delta\mathbf{p}_{i}
=
\mathbf{p}_{i}-\mathbf{p}_{i-1},
\label{eq:action_features}
\end{equation}
with $\mathbf{p}_{0}=(0,0)$.
Here, waypoint positions describe the intended path,
successive displacements capture motion over fixed time
intervals, and the normalized index indicates the
corresponding temporal position.

The action encoder combines these features with the
maneuver descriptor to obtain
\begin{equation}
\mathbf{C}_{A}
=
T_{A}
\left(
\mathbf{m},
\{\mathbf{a}_{i}\}_{i=1}^{N_a}
\right).
\label{eq:action_tokens}
\end{equation}
This representation retains individual waypoint tokens
alongside a summary of the full trajectory, allowing the
world model to use motion cues at specific horizons
together with the overall course of ego movement.
The resulting action condition guides future geometry
prediction according to the supplied ego behavior.

\textbf{Semantic Adaptation.}
While the action condition specifies how the ego vehicle
is intended to move, it does not explicitly describe
the surrounding traffic context or interactions among
road users. Similar ego trajectories may occur in
different situations, such as slowing down behind a
leading vehicle or yielding to a crossing pedestrian,
with different implications for future scene evolution.
We therefore introduce observation-grounded VLM features
to complement ego-motion conditions with scene semantics
and interaction cues, helping the world model predict
futures that are consistent with both the supplied action
and the observed driving context.

The semantic branch uses the Qwen2.5-VL
backbone~\cite{bai2025qwen25vl} of the VLA
to process the available multi-camera observations and
textual prompt. We extract hidden states $\mathbf{H}_{L}$
from layer 22. Assistant-side responses are removed before
constructing the semantic input, preventing target answers
from entering this feature-extraction path.

The hidden sequence is normalized and projected into the
world-model condition space:
$\overline{\mathbf{H}}_{L}
=\operatorname{Proj}_{L}(\operatorname{LN}(\mathbf{H}_{L}))$.
A set of 16 learnable queries $\mathbf{Q}_{L}$ aggregates
the variable-length sequence into compact semantic tokens:
\begin{equation}
\mathbf{C}_{L}
=
\operatorname{LN}
\left(
\operatorname{CrossAttn}
\left(
\mathbf{Q}_{L},
\overline{\mathbf{H}}_{L},
\overline{\mathbf{H}}_{L}
\right)
\right).
\label{eq:semantic_adapter}
\end{equation}
Padding positions are masked during cross-attention.
The resulting semantic tokens have a fixed length and
a feature dimension compatible with the world model.
VGGT geometry tokens are not directly injected into the
VLA; geometric and semantic information are combined
during future-world prediction.

\subsection{Action--Semantic 3D World Evolution}
\label{sec:conditioned_world}

Historical observations alone do not explicitly specify
which ego behavior should govern future prediction.
We therefore model
$p_{\theta}(\mathbf{Z}^{f}\mid
\mathbf{Z}^{c},\mathbf{C}_{A},\mathbf{C}_{L})$,
where action tokens provide motion conditions and semantic
tokens supply complementary driving context.
These conditions guide the evolution of the future-token
stream while historical geometry anchors the prediction
to the observed scene.

\textbf{Layer-Wise Condition Injection.}
The flow transformer contains eight dual-stream blocks
followed by eight single-stream blocks.
At each dual-stream block, the future tokens first interact
with the historical geometry through the backbone and
then receive action and semantic conditions through two
separate residual cross-attention modules.

\begin{figure*}[!t]
    \centering
    \includegraphics[width=\textwidth]
    {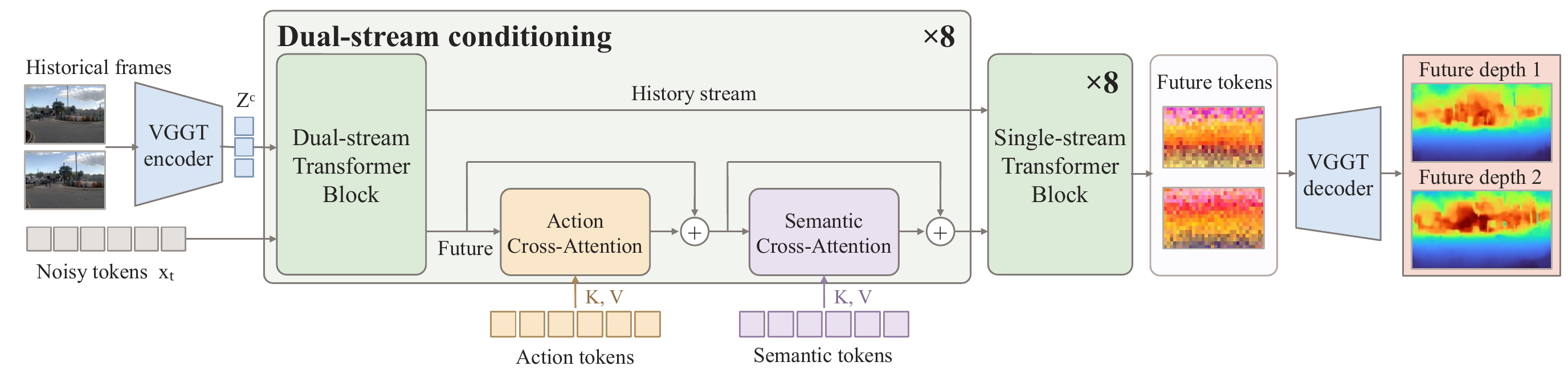}
    \caption{
        \textbf{Action--semantic 3D world evolution.}
        Each of the eight dual-stream blocks is followed by
        action and semantic cross-attention that sequentially
        update the future stream, with layer-specific parameters.
        The historical and conditioned future streams are then
        jointly processed by eight single-stream blocks to
        predict future geometry.
    }
    \label{fig:conditioned_world}
\end{figure*}

Let $\mathbf{X}_{l}^{f}$ denote the future hidden states
after the $l$-th dual-stream block. The conditional updates
are
\begin{align}
\widetilde{\mathbf{X}}_{l}^{f}
&=
\mathbf{X}_{l}^{f}
+
\operatorname{CA}_{A}^{l}
\left(
\operatorname{LN}(\mathbf{X}_{l}^{f}),
\operatorname{LN}(\mathbf{C}_{A})
\right),
\label{eq:action_injection}
\\
\overline{\mathbf{X}}_{l}^{f}
&=
\widetilde{\mathbf{X}}_{l}^{f}
+
\operatorname{CA}_{L}^{l}
\left(
\operatorname{LN}(\widetilde{\mathbf{X}}_{l}^{f}),
\operatorname{LN}(\mathbf{C}_{L})
\right).
\label{eq:semantic_injection}
\end{align}
The action and semantic modules have independent parameters,
and each layer uses its own pair of modules.
This ordering introduces ego-motion information before
semantic refinement; it does not impose a strict
disentanglement between their effects. The conditioned
states are passed to the subsequent world-model blocks
to produce future-token predictions.
Fig.~\ref{fig:conditioned_world} illustrates this
layer-wise conditioning mechanism.

\textbf{Conditional Flow Matching.}
Let $\mathbf{x}_{1}=\mathbf{Z}^{f}$ denote the target future
tokens and $\mathbf{x}_{0}\sim\mathcal{N}(\mathbf{0},\mathbf{I})$
denote Gaussian noise. We construct the interpolated state
\begin{equation}
\mathbf{x}_{t}
=
(1-t)\mathbf{x}_{0}+t\mathbf{x}_{1},
\qquad t\in[0,1].
\label{eq:flow_interpolation}
\end{equation}
The model learns a conditional velocity field through
\begin{equation}
\mathcal{L}_{\mathrm{flow}}
=
\mathbb{E}
\left[
\left\|
v_{\theta}
\left(
\mathbf{x}_{t},t;
\mathbf{Z}^{c},\mathbf{C}_{A},\mathbf{C}_{L}
\right)
-
(\mathbf{x}_{1}-\mathbf{x}_{0})
\right\|_{2}^{2}
\right].
\label{eq:flow_loss}
\end{equation}
We use an endpoint-prediction parameterization: the network
predicts future tokens
$\widehat{\mathbf{Z}}^{f}_{t}$ at the sampled flow time,
with the corresponding velocity field defined by
\begin{equation}
\widehat{\mathbf{Z}}^{f}_{t}
=
\mathbf{x}_{t}
+
(1-t)\,
v_{\theta}
\left(
\mathbf{x}_{t},t;
\mathbf{Z}^{c},\mathbf{C}_{A},\mathbf{C}_{L}
\right).
\label{eq:flow_endpoint}
\end{equation}
This formulation connects conditional token prediction
with continuous transport from noise to future geometry.

\textbf{Future Geometry Supervision.}
To preserve geometric decodability, the predicted endpoint
tokens are concatenated with the historical tokens and
processed by the pretrained VGGT second-stage aggregator
and depth head:
\begin{equation}
\widehat{\mathbf{D}}^{f}_{t}
=
\left[
D_{\mathrm{depth}}
\left(
E_{\mathrm{G}}^{\mathrm{part2}}
\left(
[\mathbf{Z}^{c},\widehat{\mathbf{Z}}^{f}_{t}]
\right)
\right)
\right]_{f}.
\label{eq:future_depth_decoding}
\end{equation}
The depth loss acts directly on the prediction at the
sampled flow time, without an additional sampling stage
during this supervision step.

A frozen VGGT teacher processes the complete real
four-frame sequence and provides future-depth pseudo-targets
$\mathbf{D}_{\mathrm{ps}}^{f}$.
We apply Smooth L1 to log-depth values:
\begin{equation}
\mathcal{L}_{\mathrm{depth}}
=
\operatorname{SmoothL1}_{\mathcal{M}}
\left(
\log\max(\widehat{\mathbf{D}}^{f}_{t},\epsilon),
\log\max(\mathbf{D}_{\mathrm{ps}}^{f},\epsilon)
\right),
\label{eq:depth_loss}
\end{equation}
where $\mathcal{M}$ selects valid finite entries and
$\epsilon$ ensures positive depth values.
The world-model objective is
\begin{equation}
\mathcal{L}_{\mathrm{world}}
=
\mathcal{L}_{\mathrm{flow}}
+
\lambda_{d}\mathcal{L}_{\mathrm{depth}}.
\label{eq:world_loss}
\end{equation}
The decoder also supports camera-pose and world-point
outputs, but these outputs do not introduce additional
pose or point losses in the described world-model
objective.

\section{Experiments}
\label{sec:experiments}

\subsection{Datasets}

We use NAVSIM~\cite{dauner2024navsim} with OpenScene, a compact redistribution
of nuPlan that provides driving images, ego states,
and trajectory annotations at 2\,Hz.
We select this setting over the datasets used in the
original VGGT-World experiments~\cite{sun2026vggtworld} because it supports
action-conditioned prediction and provides a framework
for subsequent planning evaluation.
The 0.5\,s interval between observations allows faster
ego and surrounding-agent motion to produce substantial
viewpoint shifts, object displacement, and occlusion changes,
challenging future geometry prediction.
As dense depth ground truth is unavailable for our setup,
we use pretrained VGGT predictions from actual future
images as a shared pseudo-ground-truth reference.

\subsection{Evaluation Metrics}

We evaluate future depth using AbsRel, RMSE, and
$\delta_1$, latent predictions using Token MSE and
cosine similarity, and 3D points using Point EPE
and relative Point EPE.
All predictions are compared against VGGT-derived
references constructed from real future observations.

\subsection{Main Results}

We evaluate future geometry prediction on a NAVSIM motion
subset of 3,000 clips, with 1,000 each for left turns,
straight driving, and right turns to balance maneuver
contributions.
All methods use the same clips, temporal stride 1,
and 10 flow sampling steps, with metrics averaged over clips.
Predictions are evaluated against pseudo-ground-truth
geometry generated by pretrained VGGT from actual future images.

Our primary baseline is the original VGGT-World.
To control for additional training on the target dataset,
we also continue training VGGT-World on NAVSIM using
visual conditioning alone, denoted as VGGT-World-N.
These baselines represent the original geometry predictor
and its NAVSIM-adapted counterpart.

Table~\ref{tab:main_depth_results} summarizes latent
and decoded-depth prediction quality.
Our method achieves the lowest Token MSE and RMSE,
and the highest cosine similarity and threshold accuracy.
RMSE decreases by 13.7\% relative to VGGT-World
and 15.7\% relative to VGGT-World-N, while
$\delta_1$ increases by 4.11 and 11.10 percentage
points, respectively.
AbsRel remains comparable, with a modest increase.
Thus, our method improves latent alignment, RMSE,
and threshold accuracy while largely maintaining
relative depth accuracy.

\begin{figure*}[!t]
    \centering
    \includegraphics[width=0.86\linewidth]{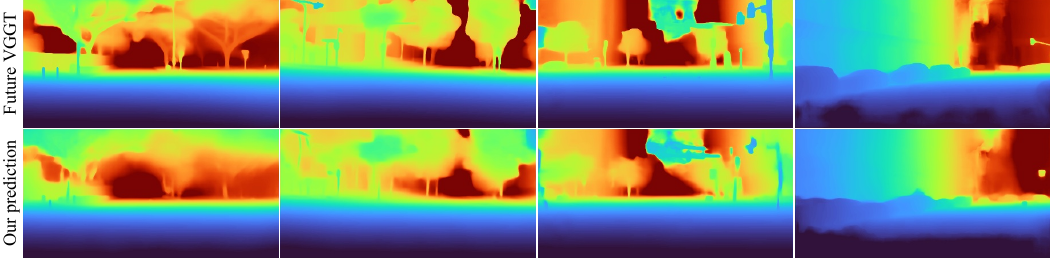}
    \caption{
    \textbf{Qualitative comparison of future depth.} Top: reference depth estimated by
    VGGT from the recorded future frames.
    Bottom: our predictions conditioned on historical
    observations, ego actions, and driving semantics.
    }
    \label{fig:future_depth_comparison}
\end{figure*}

The comparison also distinguishes latent prediction
quality from decoded geometric accuracy.
VGGT-World-N improves token-level metrics over the
original model but shows inconsistent depth improvements.
Our method further improves latent prediction, RMSE,
and threshold accuracy, while VGGT-World-N retains
the lowest AbsRel.
These findings support evaluating complementary geometric
metrics alongside latent reconstruction and motivate
the following analyses of action representation
and conditioning. Fig.~\ref{fig:future_depth_comparison} compares our
predicted future depth with pseudo-ground-truth depth
generated by VGGT from the actual future frames.

\begin{table*}[htbp]
\centering
\small
\setlength{\tabcolsep}{6pt}
\caption{Future geometry prediction on the NAVSIM
3,000-clip motion subset.
VGGT-World-N denotes visual-only continuation training
on NAVSIM. All methods use stride 1 and 10 flow sampling
steps, with depth evaluated against VGGT-derived
pseudo-ground truth.
The best and second-best results are highlighted
in bold and underlined, respectively.}
\label{tab:main_depth_results}
\begin{tabular}{lccccc}
\toprule
Method
& \shortstack{Token MSE $\downarrow$}
& Cosine $\uparrow$
& AbsRel $\downarrow$
& RMSE $\downarrow$
& $\delta_1$ $\uparrow$ \\
\midrule
VGGT-World~\cite{sun2026vggtworld}
& 0.011065
& 0.769799
& \underline{0.385980}
& \underline{0.426050}
& \underline{0.418000} \\

VGGT-World-N
& \underline{0.010620}
& \underline{0.781621}
& \textbf{0.376136}
& 0.436463
& 0.348113 \\

Ours
& \textbf{0.010459}
& \textbf{0.785120}
& 0.392457
& \textbf{0.367831}
& \textbf{0.459124} \\
\bottomrule
\end{tabular}
\end{table*}

\subsection{Ablation Studies}

We examine the roles of action conditioning and VLM-derived
semantic information in future geometry prediction.
Our analysis considers different action encodings, the
sensitivity of the trained model to action-input interventions,
and the additional role of VLM tokens. We focus on left-turn,
straight-driving, and right-turn scenarios for maneuver-specific
evaluation and visualization.

\textbf{Action Encoding.}
We compare four action representations: one-hot maneuver
encoding, trajectory positions, velocity-augmented
trajectories, and their joint representation.
These capture maneuver category, spatial path, and
temporal motion, respectively.
All configurations are evaluated on the same 3,000-clip
motion subset covering left turns, straight driving,
and right turns.

\begin{table*}[t]
\centering
\small
\setlength{\tabcolsep}{6pt}
\caption{Comparison of action-representation configurations
on the 3,000-clip motion subset.
Joint combines one-hot, position-based, and
velocity-augmented representations.
The best and second-best results are highlighted in bold
and underlined, respectively.}
\label{tab:action_encoding}
\begin{tabular}{lcccccc}
\toprule
Encoding
& Token MSE $\downarrow$
& AbsRel $\downarrow$
& $\delta_1$ $\uparrow$
& RMSE $\downarrow$
& Point EPE $\downarrow$
& Point Rel.\ EPE $\downarrow$ \\
\midrule
One-hot
& \textbf{0.01019}
& 0.42476
& \textbf{0.45279}
& \underline{0.37708}
& \underline{0.36640}
& 0.38590 \\

Trajectory
& 0.01068
& 0.40515
& 0.40943
& 0.39119
& 0.37805
& 0.38657 \\

Trajectory + velocity
& 0.01062
& \underline{0.40097}
& 0.41798
& 0.38517
& 0.37107
& \underline{0.38471} \\

Joint
& \underline{0.01050}
& \textbf{0.39320}
& \underline{0.45184}
& \textbf{0.37093}
& \textbf{0.35807}
& \textbf{0.36842} \\
\bottomrule
\end{tabular}
\end{table*}

Table~\ref{tab:action_encoding} shows that the individual
representations exhibit different strengths.
One-hot encoding achieves the lowest Token MSE and the
highest depth threshold accuracy, whereas the continuous
trajectory configurations obtain lower AbsRel.
Among the two continuous configurations, incorporating
velocity yields better results across all reported metrics
than using positions alone.

The joint configuration provides the strongest overall
decoded-geometry performance, achieving the lowest depth
and point prediction errors while retaining competitive
threshold accuracy. This pattern supports combining
action information at two levels: one-hot encoding
provides a coarse maneuver template, while trajectory
positions and velocities specify the spatial path and
motion rate within that template. The categorical
representation captures the overall maneuver, and the
continuous representations describe its detailed execution.
The stronger overall performance of the joint configuration
suggests that these two levels provide complementary
information for action-conditioned future geometry prediction.

\textbf{Action-Input Interventions.}
We assess action utilization using correct, zeroed,
shuffled, and wrong-maneuver inputs on the 3,000-clip
motion subset. Only the one-hot maneuver, trajectory
positions, and derived velocities change; model parameters,
history, semantic inputs, and initial noise remain fixed.

Table~\ref{tab:action_intervention} shows that correct
actions achieve the lowest Token MSE, AbsRel, RMSE,
and Point EPE. Shuffled and wrong-maneuver inputs degrade
all reported metrics, with wrong maneuvers causing the
largest deterioration.
These results demonstrate the model's use of scene-aligned
actions alongside visual history and semantic context.

\begin{table*}[t]
\centering
\small
\setlength{\tabcolsep}{6pt}
\caption{Action-input interventions on the 3,000-clip motion subset.
The observed history, semantic inputs, and initial sampling
noise are fixed while the action input changes.
The best and second-best results are highlighted in bold
and underlined, respectively.}
\label{tab:action_intervention}
\begin{tabular}{lccccc}
\toprule
Action input
& \shortstack{Token MSE $\downarrow$}
& AbsRel $\downarrow$
& RMSE $\downarrow$
& $\delta_1$ $\uparrow$
& \shortstack{Point EPE $\downarrow$} \\
\midrule
Correct
& \textbf{0.010459}
& \textbf{0.392457}
& \textbf{0.367831}
& \textbf{0.459124}
& \textbf{0.354330} \\

Zero
& \underline{0.010695}
& 0.414170
& \underline{0.375252}
& \underline{0.459097}
& \underline{0.356336} \\

Shuffled
& 0.010738
& \underline{0.409405}
& 0.384130
& 0.434404
& 0.377294 \\

Wrong
& 0.010913
& 0.419565
& 0.396635
& 0.412820
& 0.391264 \\
\bottomrule
\end{tabular}
\end{table*}

\begin{figure*}[t]
\centering
\includegraphics[width=0.86\textwidth]{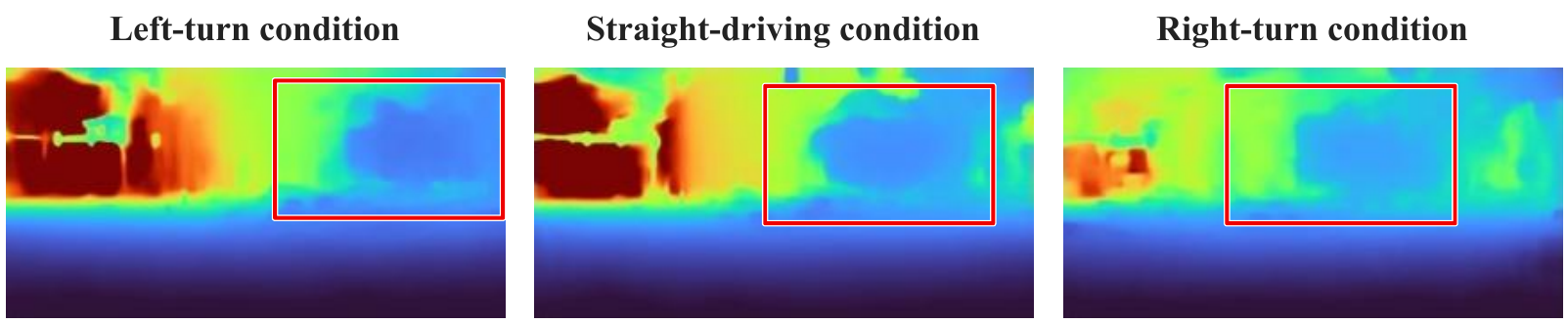}
\caption{\textbf{Action-conditioned future depth predictions} under
left-turn, straight-driving, and right-turn inputs.
Red boxes highlight corresponding image regions
with action-dependent differences in predicted geometry.}
\label{fig:action_depth_visualization}
\end{figure*}

Fig.~\ref{fig:action_depth_visualization} compares
left-turn, straight-driving, and right-turn predictions
under fixed history, semantics, and initial noise.
The highlighted regions show action-dependent viewpoint
changes, with scene structures shifting rightward under
left turns and leftward under right turns.
The corresponding changes in structural boundaries and
depth patterns provide qualitative evidence that ego
actions guide future geometry prediction.

\textbf{Action-Only vs.\ VLM--Action Conditioning.}
We compare the action-only Joint configuration, which
combines one-hot maneuvers, trajectory positions, and
velocities, with Joint + VLM, the full model used in
our main results. The latter adds semantic tokens through
an independent branch. Both are evaluated on the same
3,000-clip motion subset.

Table~\ref{tab:vlm_ablation} shows modest improvements
across all reported metrics for Joint + VLM.
The gains extend from latent alignment to decoded depth
and 3D points, suggesting that semantic conditioning
benefits the resulting geometry as well as token prediction.
While action inputs specify ego motion, VLM features
provide complementary scene context that can help
interpret its geometric consequences.
These results support combining semantic context with
action conditions for future geometry prediction.

\begin{table*}[t]
\centering
\small
\setlength{\tabcolsep}{6pt}
\caption{Comparison of Joint action conditioning and
Joint + VLM.}
\label{tab:vlm_ablation}
\begin{tabular}{lcccccc}
\toprule
Conditioning & Token MSE $\downarrow$ & AbsRel $\downarrow$ & $\delta_1$ $\uparrow$
& RMSE $\downarrow$ & Point EPE $\downarrow$ & Point Rel.\ EPE $\downarrow$ \\
\midrule
Joint & {0.010496} & {0.393195} & {0.451837} & {0.370929}
& {0.358067} & {0.368416} \\
Joint + VLM & \textbf{0.010459} & \textbf{0.392457} & \textbf{0.459124}
& \textbf{0.367831} & \textbf{0.354330} & \textbf{0.362280} \\
\bottomrule
\end{tabular}
\end{table*}

\section{Conclusion}

We present an intention-conditioned 3D world
model that integrates driving semantics and ego-motion
representations into geometry-aware future prediction.
Experiments on NAVSIM demonstrate improved latent
alignment, depth RMSE, and threshold accuracy over both
the original VGGT-World and its NAVSIM-adapted baseline,
with a modest increase in AbsRel.
The action-representation comparison supports combining
coarse maneuver templates with continuous motion details,
while action-input interventions show that correctly
aligned actions improve future geometry prediction.
The comparison with action-only conditioning shows
modest gains across all reported metrics when VLM tokens
are included.
Together, these findings support further development
of conditional geometric world models for autonomous
driving.

Effectively integrating semantic information to obtain
larger gains while preserving geometric prediction
quality remains an open challenge.
Another practical consideration is the additional
computational cost of the VLA module, which may limit
inference frequency in real-time deployment.
Future work will explore more effective semantic fusion,
lightweight conditioning, and asynchronous execution,
with semantic features updated less frequently than
action-conditioned geometry predictions.
We also plan to investigate multi-action future rollouts
and closed-loop planning evaluation, allowing candidate
actions to be assessed through their predicted geometric
consequences within an imagine--evaluate--act framework.

\bibliographystyle{IEEEtran}
\bibliography{references}

\clearpage
\appendices

\section{Implementation Details}
\label{app:implementation}
This appendix details the representation interfaces,
action and semantic feature construction, inference
configuration, and evaluation metrics.
The world predictor combines historical geometry,
supplied ego motion, and observable driving semantics
to predict future VGGT representations.

\subsection{Representation interfaces}
The geometry branch (Table~\ref{tab:app_configuration}) encodes two historical images from a single selected
camera stream. Each frame is represented by VGGT tokens with 1,024 channels.
The flow transformer operates at an internal width of 512 with eight attention
heads. It contains eight dual-stream blocks followed by eight single-stream
blocks. After each dual-stream block, separate action and semantic
cross-attention modules update the future stream in that order. Every block
has its own pair of conditioning modules. The historical and conditioned
future states subsequently enter the single-stream stage, and the output is
mapped back to the VGGT token representation.

\begin{table}[!tb]
\centering\small
\caption{Representation and inference configuration. The action-token count
below refers to the continuous trajectory component of the action interface.}
\label{tab:app_configuration}
\begin{tabular}{@{}p{0.48\linewidth}p{0.45\linewidth}@{}}
\toprule
Component & Configuration \\
\midrule
Geometry input / prediction & Two historical / two future frames \\
VGGT representation width & 1,024 \\
Flow-transformer width / attention heads & 512 / 8 \\
Dual-stream / single-stream blocks & 8 / 8 \\
Condition injection & Action CA, then semantic CA; per-layer parameters \\
Continuous trajectory & 8 ego-centric waypoints \\
Waypoint feature & Position, displacement, normalized temporal index \\
Continuous action tokens & 8 waypoint tokens and 1 summary token \\
Semantic backbone / feature layer & Driving-adapted Qwen2.5-VL / layer 22 \\
Semantic adapter & 16 queries; 8-head cross-attention \\
Inference & 10 Euler sampling steps \\
\bottomrule
\end{tabular}
\end{table}

\subsection{Action feature construction}
The Joint representation combines a one-hot maneuver descriptor with
continuous position and motion information. For the trajectory component,
let $\mathbf p_i=(x_i,y_i)$ denote an ego-centric waypoint. The coordinate
normalization used by the trajectory interface is
\begin{equation}
\begin{gathered}
\widetilde{\mathbf p}_i=(x_i/30,\;y_i/10),\\
\mathbf a_i=[\widetilde{\mathbf p}_i;
\widetilde{\mathbf p}_i-\widetilde{\mathbf p}_{i-1};i/8],
\\ \widetilde{\mathbf p}_0=\mathbf 0.
\end{gathered}
\end{equation}
The five-dimensional features are projected by a two-layer MLP with a SiLU
activation. Learned waypoint and type embeddings are added before layer
normalization. A summary token pools the waypoint tokens and adds learned
summary and type embeddings. At fixed temporal spacing, successive
displacements encode the same motion progression as velocities up to a
constant scale. The same supplied trajectory conditions both predicted frames;
its temporal ordering remains available through the individual waypoint tokens.

\subsection{Semantic feature extraction}
The semantic branch processes the multi-camera observations associated with
the conditioning timestamp and their textual prompt. Its inputs retain the
system and user messages. Assistant-side target responses are excluded before
the chat sequence is constructed. Hidden states are captured after the 22nd
language-model layer, normalized, and linearly projected to the 512-dimensional
condition space. Sixteen learned queries attend to this sequence, using the
processor attention mask to exclude padding. Output layer normalization
produces the semantic tokens. The adapter uses the attention output directly;
residual updates occur in the subsequent world-model conditioning modules.
The geometric and semantic branches therefore retain their respective input
organizations while supplying compatible conditions to the predictor.

\subsection{Evaluation Metrics}
\label{app:evaluation_metrics}

We evaluate future predictions in depth, latent-token,
and 3D point spaces.
Reference representations are generated by pretrained
VGGT from the complete real image sequence, selecting
the outputs corresponding to the future frames.
These references are shared across the compared methods
and serve as pseudo-ground truth.

\paragraph{Depth Metrics.}
We report absolute relative error (AbsRel), root mean
squared error (RMSE), and threshold accuracy ($\delta_1$).
AbsRel measures absolute depth error normalized by
the reference depth, while RMSE measures the square
root of the mean squared depth error.
Threshold accuracy is the proportion of valid pixels
satisfying
$\max(\widehat{d}/d,d/\widehat{d})<1.25$,
where $\widehat{d}$ and $d$ denote predicted and
reference depth, respectively.
Lower AbsRel and RMSE indicate smaller errors,
whereas higher $\delta_1$ indicates better agreement.

\paragraph{Latent Metrics.}
Token MSE measures the mean squared difference between
predicted future tokens and their VGGT-derived targets.
Cosine similarity measures the directional agreement
of the corresponding feature representations.
Lower Token MSE and higher cosine similarity indicate
better latent prediction quality.

\paragraph{Point Metrics.}
Point endpoint error (Point EPE) measures the Euclidean
distance between corresponding predicted and reference
3D points.
Relative Point EPE additionally expresses point error
relative to the reference geometry.
Lower values indicate better agreement.
Together, latent and decoded-geometry metrics assess
both representation prediction and its geometric outputs.

Since the references are generated by VGGT, the reported
metrics quantify agreement with the teacher rather than
direct accuracy against sensor-measured ground-truth
geometry.

\section{Temporal Organization, Training, and Sampling}
\label{app:training}
\subsection{Temporal and camera organization}
A geometry clip uses indices $(k-s,k,k+s,k+2s)$ from one camera stream,
where $k$ is the conditioning sample and $s$ is the temporal stride. The first
pair provides history and the second pair defines the future targets. At the
nominal 2\,Hz rate, stride 1 yields the offsets in Table~\ref{tab:app_temporal}.
Semantic inputs and ego actions are associated with sample $k$.

\begin{table}[!t]
\centering\footnotesize
\setlength{\tabcolsep}{2pt}
\caption{Temporal roles within a four-frame clip. Future RGB is used to
construct references; the world predictor receives historical geometry and
conditions from the conditioning sample.}
\label{tab:app_temporal}
\begin{tabular}{@{}lcccc@{}}
\toprule
& History 1 & History 2 & Future 1 & Future 2 \\
\midrule
Sample index & $k-s$ & $k$ & $k+s$ & $k+2s$ \\
Nominal offset, $s=1$ & $-0.5$ s & $0$ s & $+0.5$ s & $+1.0$ s \\
Historical encoder input & Yes & Yes & --- & --- \\
Full-sequence teacher input & Yes & Yes & Yes & Yes \\
Future target selected & --- & --- & Yes & Yes \\
\bottomrule
\end{tabular}
\end{table}

Continuous runs require matching scene and camera identities, consecutive
source indices and numerical sample identifiers, and existing image files.
The semantic branch preserves the conditioning message's three views:
\texttt{CAM\_L0}, \texttt{CAM\_F0}, and \texttt{CAM\_R0}. Thus, geometry
uses temporal observations from one camera, while semantics uses wider
spatial context at the conditioning timestamp.

\subsection{Historical encoding and teacher supervision}
Two independent encoder calls produce the historical condition and future
targets: one processes history only, and the other processes all four images
and selects future-frame tokens. A frozen full-sequence teacher also provides
future depth and point references. Generated future tokens are concatenated
with historical tokens and decoded through the second-stage aggregator and
pretrained geometry heads.

\subsection{Endpoint prediction and flow training}
With noise level $\sigma=1-t$, future-token target $\mathbf x_1$, and
Gaussian noise $\mathbf x_0\sim\mathcal N(\mathbf 0,\mathbf I)$, training uses
\begin{equation}
\mathbf x_\sigma=\sigma\mathbf x_0+(1-\sigma)\mathbf x_1,\qquad
\widehat{\mathbf v}_\sigma=
\frac{\widehat{\mathbf x}_1-\mathbf x_\sigma}{\sigma}.
\end{equation}
The endpoint $\widehat{\mathbf x}_1$ is predicted under historical, action,
and semantic conditions. The target velocity is
$(\mathbf x_1-\mathbf x_\sigma)/\sigma$. The implementation samples from
1,000 discrete noise levels and minimizes the MSE between these velocities.
The depth branch decodes this same endpoint at the sampled noise level.
Positive-clamped depths, with $\epsilon=10^{-3}$, enter a Smooth L1 loss in
log space. The world objective combines flow loss with depth loss weighted
by $0.1$. AdamW is used with weight decay $0.05$ for the world-predictor
parameter group and gradient clipping at $1.0$; conditioning groups are
optimized separately. Computation uses BF16 where supported.

\subsection{Euler inference}
Starting from Gaussian future tokens, sampling keeps the conditions fixed
and updates the state over descending noise levels:
\begin{equation}
\mathbf x_{\sigma_{j+1}}=\mathbf x_{\sigma_j}+
\frac{\sigma_j-\sigma_{j+1}}{\sigma_j}
\left(\widehat{\mathbf x}_{1,j}-\mathbf x_{\sigma_j}\right).
\end{equation}
After ten updates, the resulting future tokens are decoded into geometry.
For action interventions, each comparison reuses the initial noise tensor,
so differences arise from the supplied action condition.

\section{Evaluation Protocol and Additional Measurements}
\label{app:evaluation}
\subsection{Token and decoded-geometry metrics}
Evaluation uses the same 3,000-clip motion subset spanning left turns,
straight driving, and right turns. Predictions are paired with their
corresponding VGGT-derived references. Let $\widehat{\mathbf z}_q$ and
$\mathbf z_q$ denote predicted and reference token vectors, with channel
count $C$ and token count $Q$ across the future frames. We compute
\begin{align}
\mathrm{Token\ MSE}&=\frac{1}{QC}\sum_{q=1}^{Q}
\|\widehat{\mathbf z}_q-\mathbf z_q\|_2^2,\\
\mathrm{Cosine}&=\frac{1}{Q}\sum_{q=1}^{Q}
\frac{\widehat{\mathbf z}_q^{\mathsf T}\mathbf z_q}
{\|\widehat{\mathbf z}_q\|_2\|\mathbf z_q\|_2},
\end{align}
with numerical stabilization for the cosine denominator. For valid positive
finite depth pairs $(\widehat d_u,d_u)$, the decoded-depth metrics are
\begin{align}
\mathrm{AbsRel}&=\frac{1}{|\mathcal V|}\sum_{u\in\mathcal V}
\frac{|\widehat d_u-d_u|}{d_u},\notag\\
\mathrm{RMSE}&=\sqrt{\frac{1}{|\mathcal V|}\sum_{u\in\mathcal V}
(\widehat d_u-d_u)^2},\\
\delta_1&=\frac{1}{|\mathcal V|}\sum_{u\in\mathcal V}
\mathbf 1\!\left[\max\!\left(\frac{\widehat d_u}{d_u},
\frac{d_u}{\widehat d_u}\right)<1.25\right].
\end{align}
Depth scores are computed per future frame, averaged over the two future
frames within a clip, and then averaged over clips. The metric functions
compare predicted and reference depths directly. For point outputs,
finite three-dimensional correspondences define the valid set $\mathcal P$:
\begin{align}
\mathrm{EPE}&=\frac{1}{|\mathcal P|}\sum_{u\in\mathcal P}
\|\widehat{\mathbf P}_u-\mathbf P_u\|_2,\\
\mathrm{Rel.\ EPE}&=\frac{1}{|\mathcal P|}\sum_{u\in\mathcal P}
\frac{\|\widehat{\mathbf P}_u-\mathbf P_u\|_2}
{\max(\|\mathbf P_u\|_2,10^{-3})}.
\end{align}
Point errors are expressed in the coordinate scale of the VGGT-derived
reference. Token, depth, and point scores describe complementary aspects
of the same future prediction.

\subsection{Controlled action inputs}
Correct uses the clip's supplied action. Zero suppresses the one-hot,
position, and velocity inputs. Shuffled assigns action inputs from another
clip, and Wrong assigns an alternative maneuver. For donor-based conditions,
the action components are transferred together to retain their internal
correspondence. The trained parameters, historical images, semantic inputs,
and initial sampling noise are unchanged. The VLM continues to receive the
current clip's observable input. All conditions are evaluated against the
same recorded future reference.

\subsection{Supplementary representation measurements}
Table~\ref{tab:app_extra} supplies measurements beyond the compact main-text
comparison. Joint combines the three action representations; Joint + VLM
adds the semantic branch. Both use the same training budget and evaluation
protocol. The cosine increase accompanies a reduction in spatially restricted
token error, complementing the decoded-depth and point results in the main text.

\begin{table}[!tb]
\centering\small
\caption{Additional latent measurements for the Joint and Joint + VLM
configurations. The near-field measurement follows the reported evaluation
mask. Bold indicates the better value within each row.}
\label{tab:app_extra}
\begin{tabular}{@{}lrr@{}}
\toprule
Metric & Joint & Joint + VLM \\
\midrule
Token cosine $\uparrow$ & 0.784190 & \textbf{0.785120} \\
Near-field Token MSE $\downarrow$ & 0.009563 & \textbf{0.009535} \\
\bottomrule
\end{tabular}
\end{table}

\section{Additional Qualitative Results}
\label{app:qualitative}
\subsection{Future tokens and decoded depth}
\begin{figure}[!tb]
\centering
\includegraphics[width=0.88\linewidth]{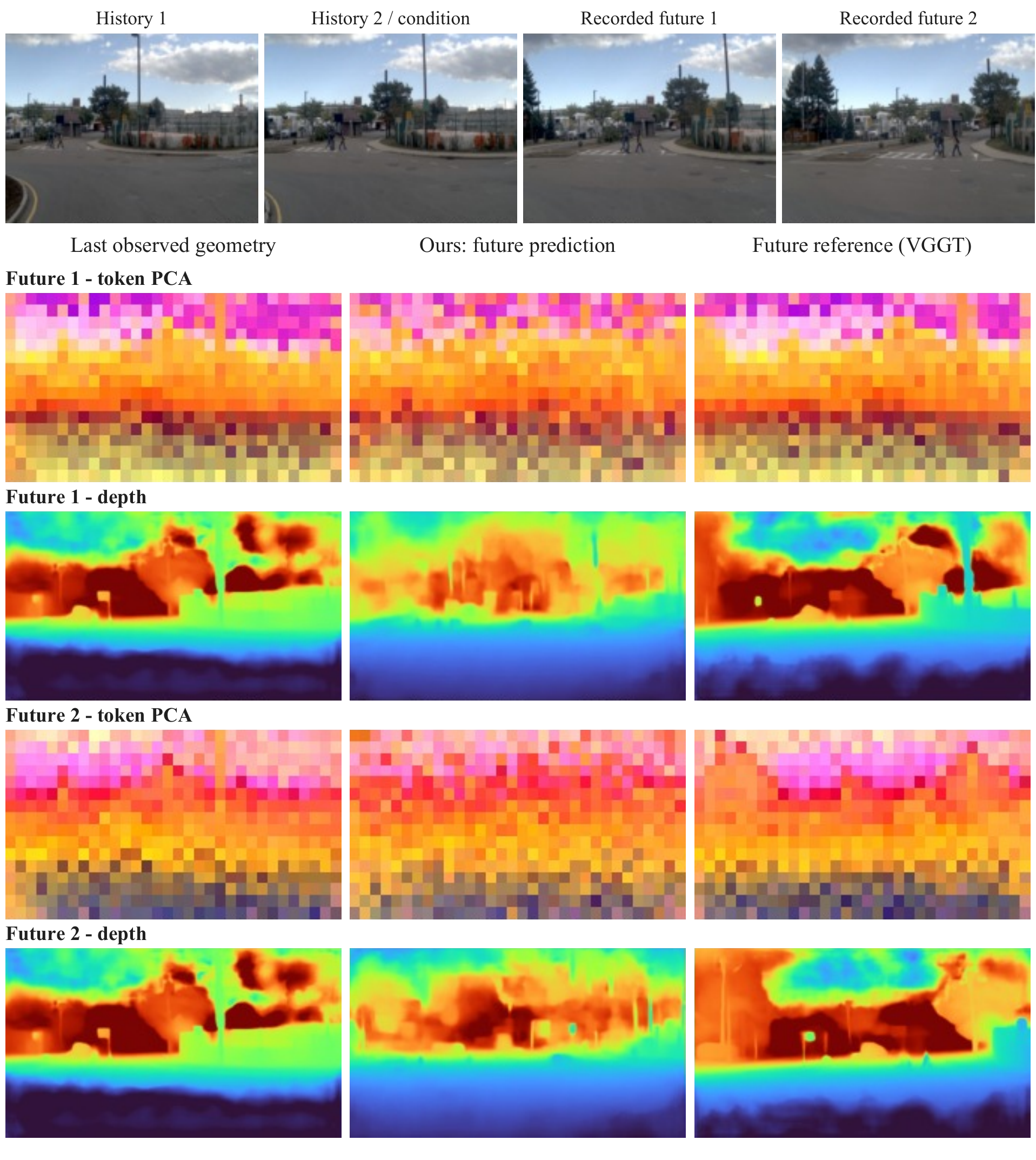}
\caption{From observed geometry to predicted future geometry.
The top row shows the recorded RGB sequence. Below, the left column
shows geometry from the last observed frame, the middle column shows
our future predictions, and the right column shows the corresponding
future reference obtained by applying VGGT to the recorded sequence.
Rows display token PCA and depth for Future 1, followed by token PCA
and depth for Future 2. The last observed geometry is reused as the
historical reference at both horizons. This arrangement shows how the
predicted representation evolves from the observed scene toward its
recorded future. Original image content and color mappings are preserved.}
\label{fig:app_latent_depth}
\end{figure}

Fig.~\ref{fig:app_latent_depth} compares the last observed geometry,
our predicted future geometry, and the corresponding VGGT-derived future
reference. The RGB strip shows two historical observations followed by two
recorded future images. Below it, three columns preserve this temporal
comparison, with token PCA and depth displayed separately at each future
horizon. Only the middle column contains our future predictions.

\subsection{Action-conditioned prediction in a curved-road scene}
\begin{figure}[!tb]
\centering
\includegraphics[width=\linewidth]{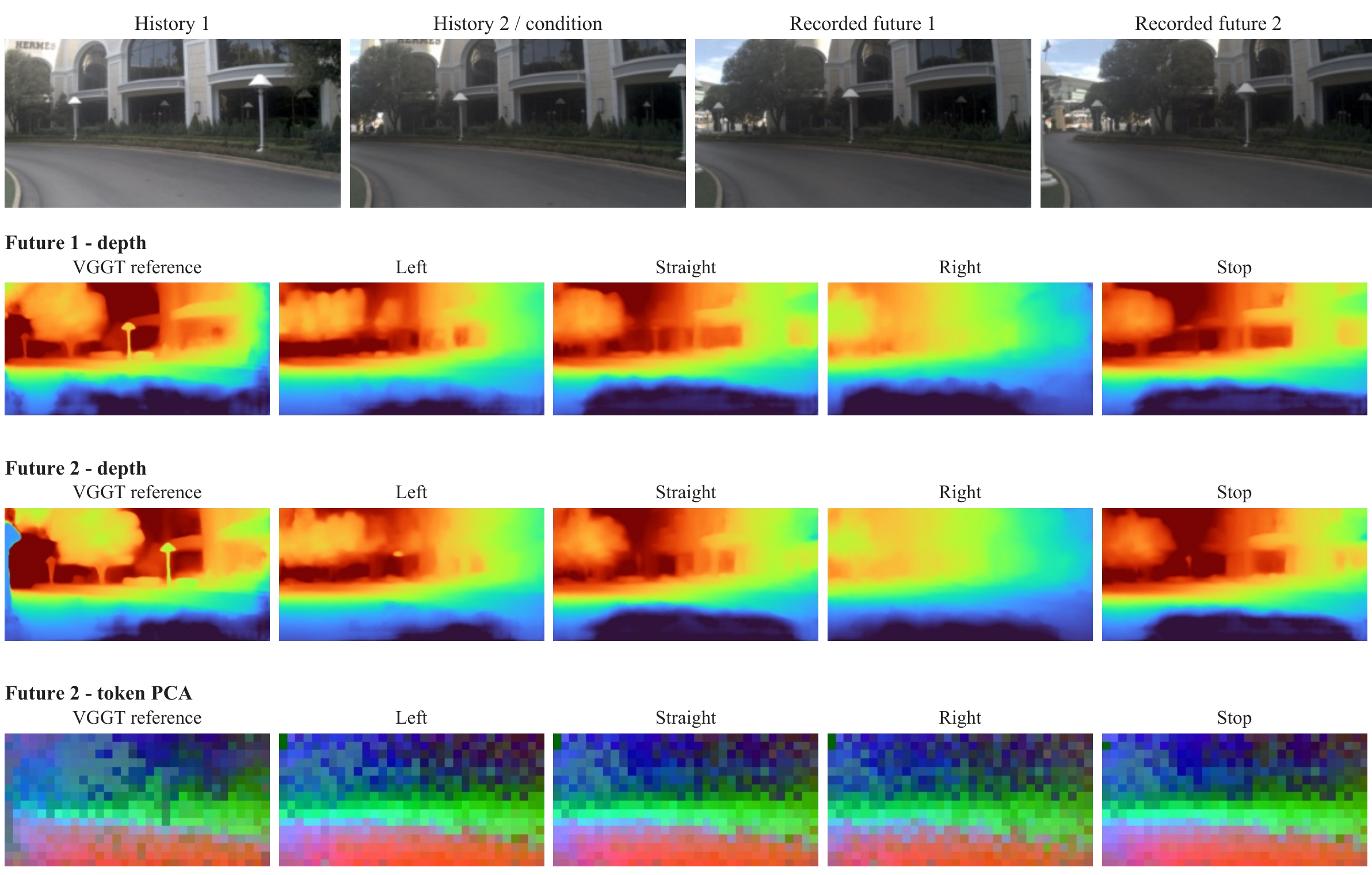}
\caption{Action-conditioned depth and token visualization in a curved-road
scene. The first row shows the recorded four-frame sequence. Subsequent rows
show first-future depth, second-future depth, and second-future token PCA.
Columns are ordered as the VGGT reference, left, straight, right, and stop.
The reference corresponds to the recorded left-turn sequence; the remaining
columns show predictions under the labeled conditions with shared history
and Euler noise. The reference is repeated conceptually across conditions
as an anchor for the recorded sequence, rather than a separate observed
future for each alternative action.}
\label{fig:app_building}
\end{figure}

Fig.~\ref{fig:app_building} provides a supplementary action-control example
with a building frontage and a curved road boundary. The recorded maneuver
is a left turn. The history is shared across the prediction columns, and the
source evaluation uses shared Euler noise. Alongside left, straight, and right
conditions, this diagnostic example includes a stop condition. The additional
condition broadens the visual comparison while retaining the same scene.

\paragraph{Action-dependent spatial organization.}
The depth panels expose changes around the building frontage, vegetation,
and road boundary. In the right-conditioned column, the arrangement of
middle-distance structures differs visibly from the left-conditioned
prediction. Straight and stop conditions offer additional comparisons of
how the same observed scene is represented under different supplied actions.
The second future row makes the temporal evolution of these spatial patterns
available for inspection.

\paragraph{Connecting latent and decoded responses.}
The final row displays the second-frame token PCA maps in the same action
order as the depth rows. Keeping both views together allows the response
of the latent world state to be examined alongside the geometry produced by
the decoder. The comparison concerns action-dependent future observations:
changes in the ego viewpoint alter the visible arrangement of otherwise
shared scene structures.

\paragraph{Relation to the quantitative intervention.}
The main-text intervention measures agreement with the recorded future when
a supplied action is retained, suppressed, or replaced. The visual comparison
here complements that measurement by displaying several action-conditioned
outputs side by side. In both cases, the observed history provides a common
geometric anchor for the future predictions.

\subsection{Additional urban-scene examples}
\begin{figure}[!tb]
\centering
\includegraphics[width=0.93\linewidth]{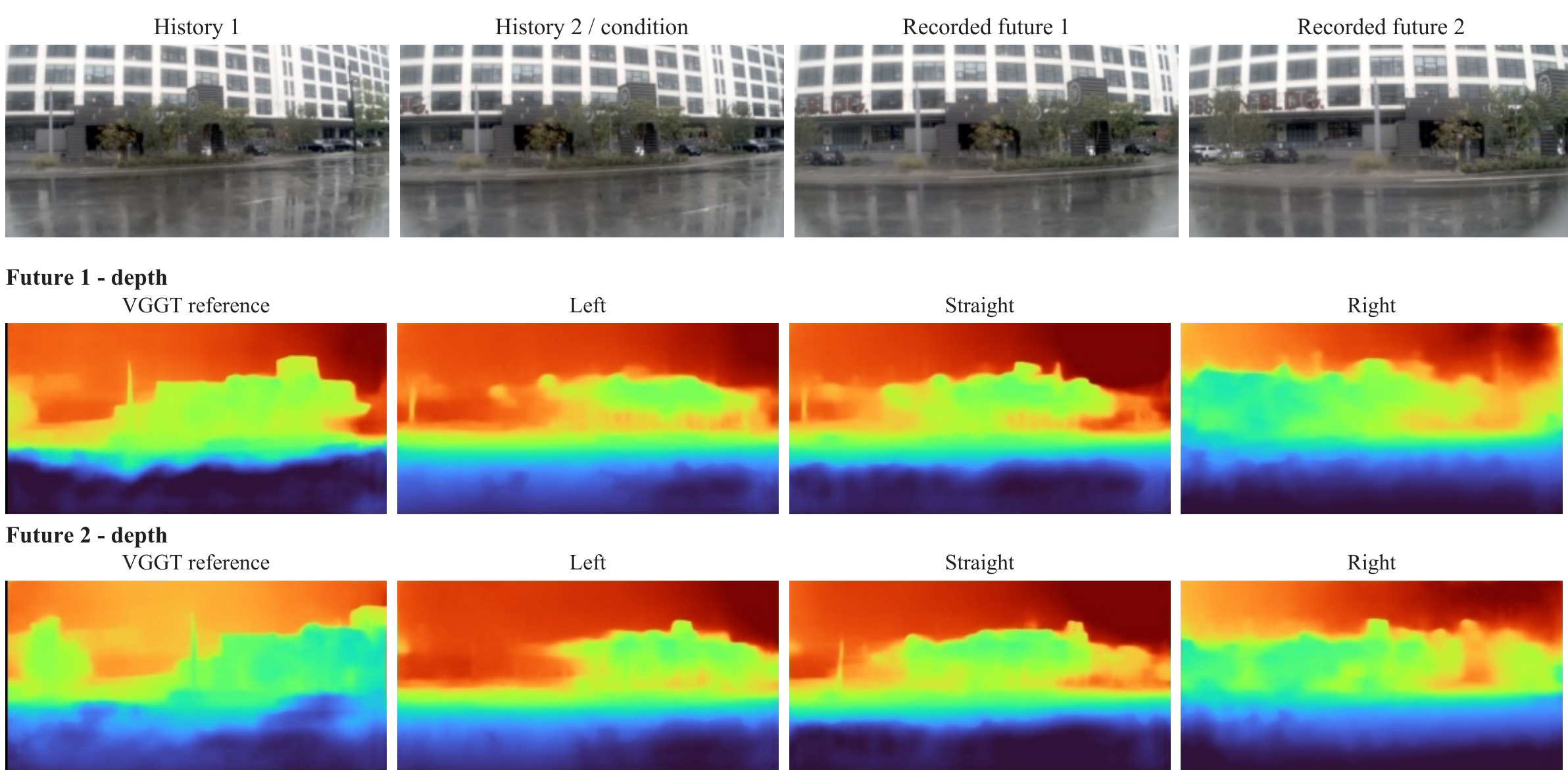}
\caption{Building-facade example with a recorded left turn. The RGB strip
provides temporal context; the depth rows compare the VGGT reference with
left-, straight-, and right-conditioned predictions at the first and second
future frames. The facade and roadside region provide spatial anchors for
inspecting changes in projected structural boundaries. History and Euler
noise are shared across the action-conditioned predictions.}
\label{fig:app_facade}
\end{figure}

\begin{figure}[!tb]
\centering
\includegraphics[width=0.93\linewidth]{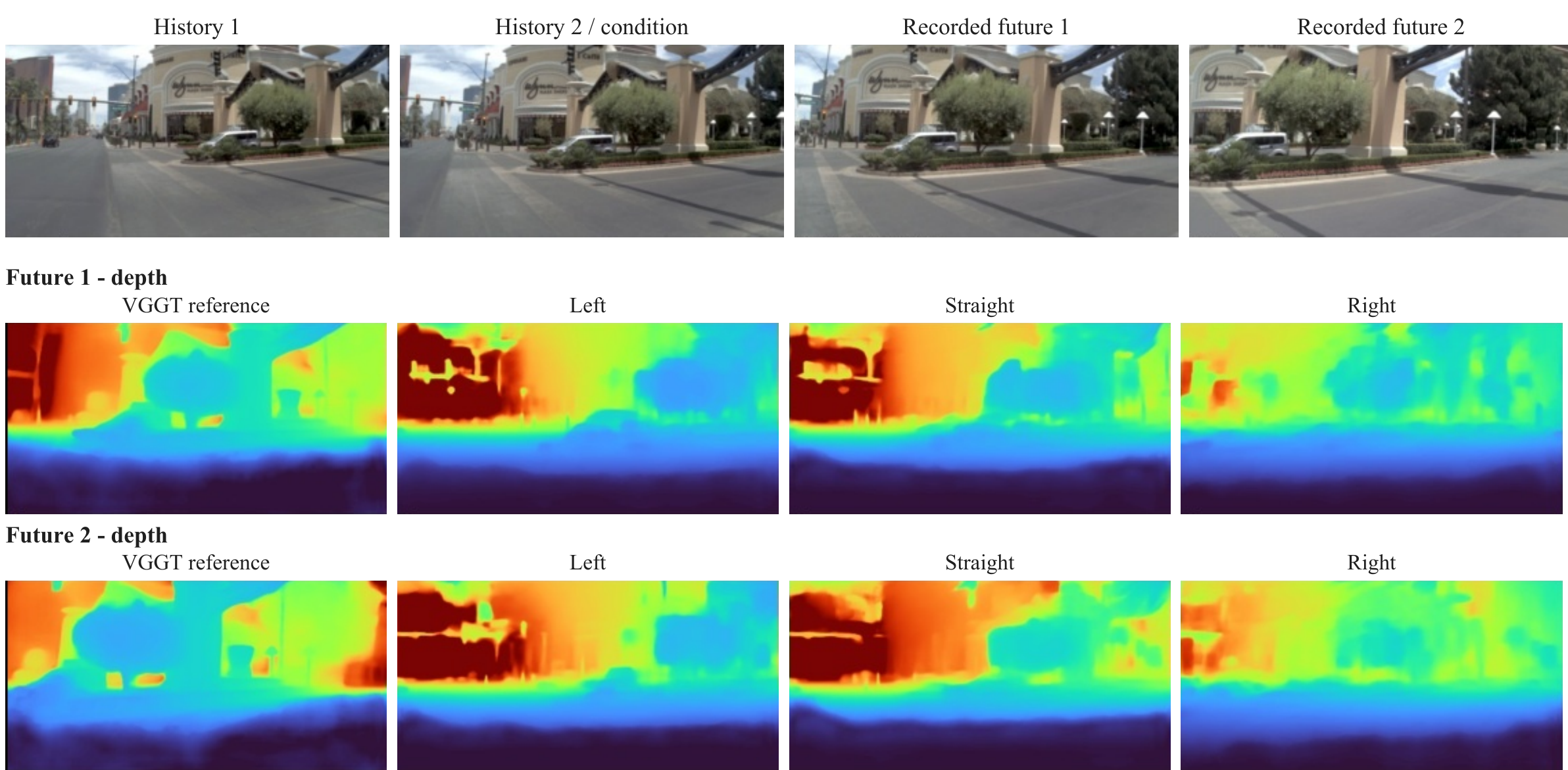}
\caption{Intersection example with a recorded right turn. The right-conditioned
second-future prediction follows the prominent central roadside structure
visible in the reference, while alternative conditions produce different
spatial depth patterns. All RGB panels are recorded frames. Reference depth
is obtained from the recorded sequence, and predictions share the historical
input and Euler noise.}
\label{fig:app_intersection}
\end{figure}

Figs.~\ref{fig:app_facade} and~\ref{fig:app_intersection} extend the visual
comparison to a broad building facade and an intersection. Both use a common
column order and show two future horizons, enabling the action-dependent
depth patterns to be compared with the recorded RGB sequence.

\end{document}